\documentclass{article}

\usepackage{microtype}
\usepackage{graphicx}
\usepackage{subfigure}
\usepackage{booktabs} %
\usepackage[utf8]{inputenc}
\usepackage[T1]{fontenc}
\usepackage{rebuttal}
\usepackage{float}
\usepackage{enumitem}

\usepackage{hyperref}

\usepackage[accepted]{icml2024}

\usepackage{coproc-macros}

\begingroup
\makeatletter
\global\let\tmp\Notice@String
\gdef\Notice@String{
    \vspace*{-3ex}\par
    \noindent
    We make our code public at \url{https://github.com/michelllepan/neural-coprocessors}. 
    \medskip\\
    \tmp
}
\makeatother
\endgroup

\usepackage[capitalize,noabbrev]{cleveref}

\usepackage[textsize=tiny]{todonotes}

\icmltitlerunning{\textbf{Coprocessor Actor Critic: Model-Based RL For Adaptive Brain Stimulation}}

\begin{document}

\twocolumn[
	\icmltitle{\texorpdfstring
		{Coprocessor Actor Critic: A Model-Based Reinforcement\\Learning Approach For Adaptive Brain Stimulation}
		{Coprocessor Actor Critic: A Model-Based Reinforcement Learning Approach For Adaptive Brain Stimulation}}

	\icmlsetsymbol{equal}{*}

	\def\argmax{\mathop{\arg\max}}
	\def\argmin{\mathop{\arg\min}}

	\begin{icmlauthorlist}
		\icmlauthor{Michelle Pan}{equal,ucb}
		\icmlauthor{Mariah Schrum}{equal,ucb}
		\icmlauthor{Vivek Myers}{ucb}
		\icmlauthor{Erdem B\i{}y\i{}k}{usc}
		\icmlauthor{Anca Dragan}{ucb}
	\end{icmlauthorlist}

	\icmlaffiliation{ucb}{Department of Electrical Engineering and Computer Sciences, UC Berkeley}
	\icmlaffiliation{usc}{Department of Computer Science, University of Southern California}

	\icmlcorrespondingauthor{Michelle Pan}{michellepan@berkeley.edu}

	\icmlkeywords{Machine Learning, ICML}

	\vskip 0.3in
]

\printAffiliationsAndNotice{\icmlEqualContribution} %

\begin{abstract}

	Adaptive brain stimulation can treat neurological conditions such as Parkinson's disease and post-stroke motor deficits by influencing abnormal neural activity.
	Because of patient heterogeneity, each patient requires a unique stimulation policy to achieve optimal neural responses.
	Model-free reinforcement learning (MFRL) holds promise in learning effective policies for a variety of similar control tasks, but is limited in domains like brain stimulation by a need for numerous costly environment interactions.
	In this work we introduce Coprocessor Actor Critic, a novel, model-based reinforcement learning (MBRL) approach for learning neural coprocessor policies for brain stimulation. %
	Our key insight is that coprocessor policy learning is a combination of learning how to act optimally in the world and learning how to induce optimal actions in the world through stimulation of an injured brain.
	We show that our approach overcomes the limitations of traditional MFRL methods in terms of sample efficiency and task success and outperforms baseline MBRL approaches in a neurologically realistic model of an injured brain.

\end{abstract}

\section{Introduction}

A neural coprocessor is a form of brain-computer interface (BCI) that can transmit signals to and from the brain \cite{Rao2019,Oehrn2023}. These interfaces can be used to treat a variety of neurological conditions by influencing abnormal neural activity \cite{Lozano2019,littleAdaptiveDeepBrain2013}.
In patients who suffer from conditions such as Parkinson's disease and dystonia, brain stimulation has the ability to steer neural activity towards activity regions which manifest in reduced disease symptoms \cite{hu2014deep,Groiss2009}. Adaptive brain stimulation can be employed not just to guide neural activity towards specific activity patterns but can also aid impaired patients in accomplishing external task objectives \cite{bryanNeuralCoprocessorsRestoring2023,dacunhaSophisticatedBasalGanglia2015}. Stroke patients are one patient population that can benefit from this aspect of brain stimulation. Stroke patients suffer injury to the brain that often results in loss of motor control and an inability to complete basic tasks, such as reaching for and grasping an object \cite{Hatem2016}. Stroke patients often struggle with these seemingly simple motor tasks due to stroke-induced lesions in the brain that can interrupt the propagation of neurological signals within and between cortical modules \cite{Ingwersen2021,Choi2023}. %
Due to the resultant motor deficits, a patient may recognize the target location and intend to move their arm to the perceived position, but struggle to do so \cite{Choi2023}. Adaptive brain stimulation exhibits potential for reducing motor impairment and restoring lost function via adaptive stimulation, enabling these patients to operate more effectively in the world \cite{ELIAS20183,gangulyModulationNeuralCofiring2022}.

\begin{figure*}[htb]
	\centering
	\includegraphics[width=\textwidth]{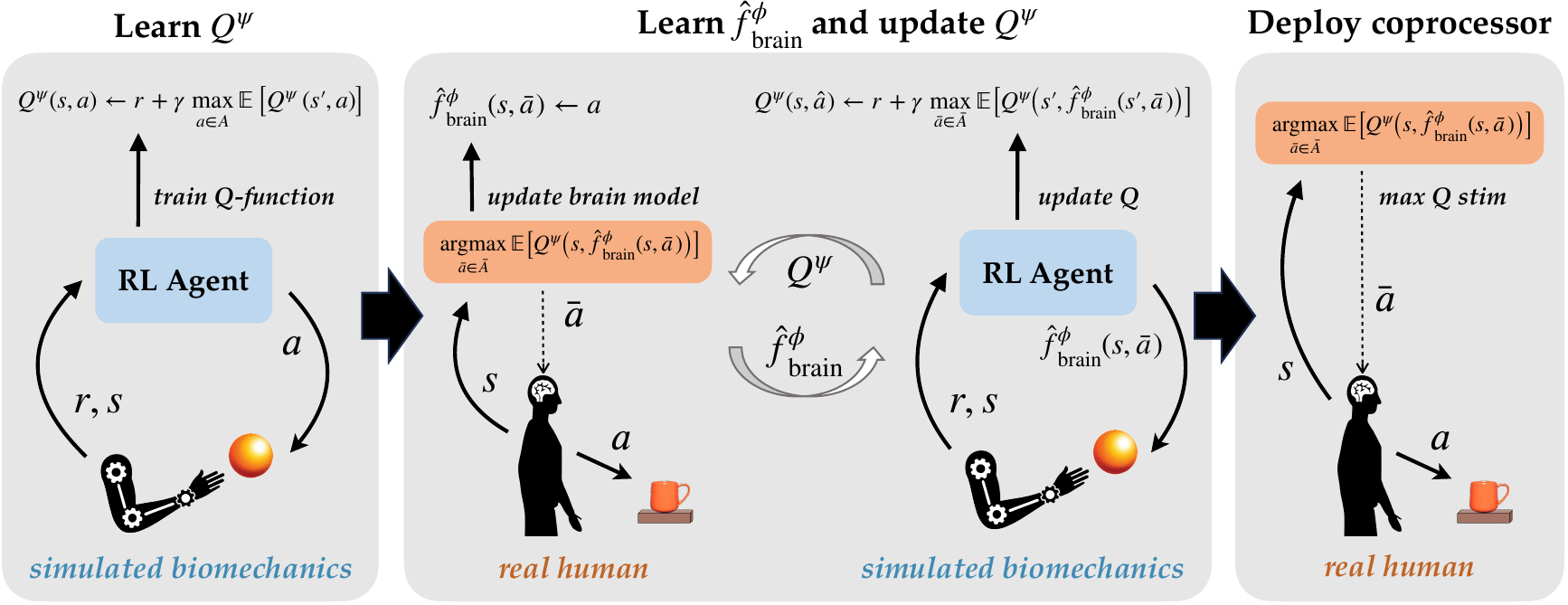}
	\caption{Overview of our framework. We first learn the Q-function, $\critic$, for world actions, $a$, via a biomechanically realistic simulator. We then learn the mapping, $\convhat$, from coprocessor actions, $\ac$, to world actions. Simultaneously, we update $\critic$ to account for the altered MDP.}
	\label{fig:overview}
\end{figure*}

In this work, we investigate the development of neural coprocessors to deliver adaptive brain stimulation for rehabilitation using an \textit{in silico} model of brain injury. Neural coprocessors rely on artificial intelligence techniques to learn a brain stimulation policy that appropriately shapes neural activity based upon the current state of the patient \cite{Rao2019}. An effective coprocessor policy can compensate for lost mobility and paretic motor deficits post-stroke \cite{bryanNeuralCoprocessorsRestoring2023}. However, there are several challenges in developing an effective coprocessor policy. Because of the heterogeneity of patients' brains, their disease manifestation, and the location of the stimuli, the optimal coprocessor policy is unique for each patient %
\cite{Visanji2022}. Furthermore, due to the complexity of the brain, closed-loop coprocessor policies are difficult to hand-engineer \cite{Oehrn2023}.  Instead, individualized coprocessor policies may be learned through interaction with the patient to ensure that the coprocessor aligns seamlessly with the unique characteristics of the patient's brain and condition. Due to the necessity for online learning, adaptive brain stimulation poses a compelling paradigm which can benefit from the application of reinforcement learning (RL) techniques.

Model-free reinforcement learning (MFRL) has shown promise for learning high-quality policies in many different control tasks that resemble brain stimulation \cite{MFRL}. However, MFRL often requires numerous environment interactions to learn a sufficiently good policy. %
When stimulating the brain, interactions with patients are costly, and thus, this domain requires an approach which can quickly learn a policy with few patient interactions. Moreover, inappropriate stimulation can produce negative side-effects such as cognitive disturbances, dyskinesia, and mood changes, further necessitating efficient learning algorithms \cite{Ashmaig2018a,grasp}. To this end, we introduce a novel model-based reinforcement learning (MBRL) approach for coprocessor stimulation that outperforms state-of-the-art MFRL and MBRL approaches in terms of both sample efficiency and task success.

Our key insight is that we can minimize online patient interaction by breaking coprocessor policy learning into two phases: 1) learning how to act optimally in the world, and 2) learning how to achieve optimal world actions via brain stimulation. With access to a biomechanically realistic simulator, we can learn the former without any patient interaction, enabling us to focus online interactions on learning the mapping from coprocessor stimulations to world actions. By separating these two components, we are able to improve the sample efficiency and performance of the coprocessor. An overview of our approach is presented in \cref{fig:overview}. We define world actions as the tangible movements performed by the patient in their environment, distinguishing them from actions (i.e., stimulations) initiated by the coprocessor.

To learn how to act optimally in the world, we rely on a biomechanical simulator. Various physics based simulators (e.g., \citet{wang2022myosim} and \citet{opensim}) have been developed that enable the physiologically realistic simulation of human biomechanics. These simulations are capable of modeling both complex human physiology and dynamic environment interactions, such as the complexities encountered in our example reaching task. We leverage a biomechanical simulator to learn the value of a human executing a world action (e.g., moving their arm) in a given state (e.g., arm joint positions), for a given task objective (e.g., reaching a target location). That is, we employ a simulator to learn the  Q-function for world actions.

While the mechanisms involved in executing a world action will be similar for every individual and can therefore be reliably modeled in simulation, because of the complexity and heterogeneity of the human brain, the effect of coprocessor stimulation cannot be as readily predicted. Instead, we must learn how to achieve optimal world actions via stimulation in an online fashion. %
To efficiently learn how coprocessor stimulations map onto world actions, we leverage the simulator-derived Q-function to guide stimulation sampling during online learning, thus focusing model learning on high-value regions of the world-action space. Because not all world actions will necessarily be realizable by the injured brain, the learned Q-function may be over-optimistic. To solve this problem,  we iteratively update the Q-function as we learn the brain model. We demonstrate that this method is orders of magnitude more efficient than learning a stimulation policy from scratch via standard MFRL. Once this mapping is learned, our coprocessor policy selects stimulations that produce world actions of maximum value, as defined by the simulator-derived Q-function.

We evaluate our approach in standard continuous control tasks as well as a novel, physiologically and neurologically realistic stroke  domain. We demonstrate that  our approach learns a better policy in fewer interactions compared to baseline MFRL and MBRL approaches. Additionally, we show that our approach is able to better aid the patient in accomplishing a given task during the online learning phase compared to baselines.
With this work, we hope to set the groundwork for implementing RL solutions for adaptive brain stimulation and to pave the way for RL researchers to further study this problem. We contribute the following:
\begin{enumerate}
	\item A novel, model-based RL approach for learning a neural coprocessor policy for closed-loop adaptive brain stimulation.
	\item A  physiologically and neurologically realistic RL benchmark environment for adaptive brain stimulation for stroke-relevant tasks.
	\item Results showing improved sample efficiency and higher training and evaluation reward relative to MFRL and MBRL baselines.
\end{enumerate}

\section{Related Work}

Current clinical applications of brain stimulation are open loop, i.e., non-adaptive \cite{khannaLowfrequencyStimulationEnhances2021,gangulyModulationNeuralCofiring2022,luApplicationReinforcementLearning2020a}. For instance, when selecting deep brain stimulation (DBS) parameters for treating Parkinson's in clinical settings, the surgeon performs a trial and error search to determine the set of stimulation parameters that best treat patient symptoms \cite{Ghasemi2018}. However, propelled by advances in medical technology and a better understanding of neurophysiological signals, closed-loop coprocessors are emerging as a new and promising paradigm \cite{Frey2022}. Closed-loop DBS relies on feedback signals either in the form of decoded brain readings or other external state information to dynamically adjust stimulation and deliver more precise interventions for patients. One challenge in closed-loop brain stimulation is the formulation of an effective control policy. Various methods have been proposed to develop adaptive DBS policies based on patient state information \cite{Little2013,Stewart,Oehrn2023}. For instance, \citet{Little2013} proposed an approach that modulates stimulation parameters based upon a user-defined threshold of local evoked potentials. However, these manually crafted strategies often fall short of capturing the intricate interplay between DBS and the brain, and are difficult to personalize for individual patients.

\paragraph{Reinforcement Learning for Brain Stimulation:} An alternative to hand-engineering policies is to leverage reinforcement learning techniques and learn a stimulation policy via data collected through patient interactions \cite{SafeMetal,Gao,Coventry}. Several prior works have explored RL methods for closed-loop brain stimulation \cite{Lu, Gao}. \citet{Gao} investigated employing a MFRL approach with offline warm-starting to learn an effective stimulation policy for Parkinson's patients. Similarly, \citet{Lu} proposed an actor-critic  method and demonstrated its performance in simulation. However, such MFRL methods require a large number of patient interactions and thus, while effective in simulation, MFRL is difficult to deploy in the real world \cite{arnold}.

\paragraph{Model Based Reinforcement Learning:} An alternative to MFRL is Model-Based Reinforcement Learning (MBRL). MBRL reduces training time and improves learning efficiency by utilizing a predictive dynamics model to learn an effective policy \cite{valencia,janner2019when}. MBRL can typically be broken down into two steps: 1) dynamic model learning followed by 2) integration and planning \cite{moerland2023modelbased}.  The downside of MBRL is that the model used for planning may be inaccurate, thus producing suboptimal plans \cite{MBRLinaccurate}. Our approach minimizes this risk by leveraging a simulation-derived Q-function to guide model learning and sample stimulations that produce high-value environment actions, enabling us to robustly and efficiently learn the brain dynamics model.

\section{Problem Setup}
We consider a world MDP with continuous state and action spaces defined by the tuple ($\S$, $\A$,  $P$, $R$). $\mathcal{S}$ defines the state space and  $\A$ the world action space. $P$: $\S \times \A \to \Delta\S$ denotes the probability distribution of the next state when action $a$ is taken at state $s$. $R :\S\times\A\to\R$, defines the reward function.

Using this MDP to describe the world, we construct a second MDP from the perspective of the coprocessor stimulations. We assume there is some (potentially nondeterministic) mapping $\conv: S\times\Ac\to \A$ that converts stimulations to the resulting world actions in a state-dependent fashion. We then define the augmented coprocessor MDP, $\Mbar= (\S, \Ac, \Pbar, R)$ where $\Ac$ is the space of possible coprocessor stimulation actions and the probability distribution over the next state is defined in \cref{eq:mbar}.
\begin{equation}
	\Pbar(s'\mid s,\ac)=\ex[][\Big]{P\parens[\big]{s'\mid s,\conv(s,\ac)}}.
	\label{eq:mbar}
\end{equation}

In \cref{eq:mbar} the expectation is over the stochasticity of $\conv$.

The objective is to learn a policy $\pi(\act\mid\st)$ that maximizes the task reward in the coprocessor MDP, $\Mbar$. In simulation we do not have access to the true brain model $\conv$, so we use a learned model $\convhat$, yielding a simulated version of the coprocessor MDP, $\Mhat=(\S,\Ac,\Phat,R)$:
\begin{equation}
	\Phat(s'\mid s,\ac)=\ex[][\Big]{P\parens[\big]{s'\mid s,\convhat(s,\ac)}}.
	\label{eq:mhat}
\end{equation}

\section{Methodology}

\begin{algorithm}
	\begin{algorithmic}[1]
		\caption{Coprocessor Actor Critic (\algname)} \label{alg:overview}
		\REQUIRE world MDP $M=(\S,\A,P,R)$, $s_0\sim p_0$
		\REQUIRE stimulation space $\Ac$, injured brain $\conv$
		\STATE initialize $\convhat$

		\WHILE {access to simulator} \label{lst:line:begincriticlearning}
		\STATE rollout $\pi$ in $M$ with experiences $(s, a, r, s')$ \label{lst:line:rolloutworldpolicy}
		\STATE fit $\critic(s,a)$ with \cref{eq:q-update} \label{lst:line:updatecritic}%
		\ENDWHILE \label{lst:line:endcriticlearning}

		\WHILE {access to injured brain} \label{lst:line:beginflearning}
		\STATE rollout $\pic$ in $\Mbar$ with experiences $(s, \ac, r, s')$\label{lst:line:rolloutcoprocpolicy}
		\STATE $a\gets\conv(s,\ac)$ \label{lst:line:getcoprocaction}
		\STATE fit $\convhat(s,\ac)$ to $a$ \label{lst:line:updatedcritic}
		\WHILE {not converged} \label{lst:line:begincriticupdates}
		\STATE rollout $\pic$ in $\Mhat$ with experiences $(s, \ac, r, s')$ \label{lst:line:rolloutcoprocinMhat}
		\STATE $\hat a\gets\convhat(s,\ac)$ \label{lst:line:getcoprocaction2}
		\STATE fit $\critic(s,\hat a)$ using \cref{eq:q-calibrate} \label{lst:line:updatecritic2}
		\ENDWHILE
		\ENDWHILE
		\STATE \textbf{return} $\pic$
	\end{algorithmic}

\end{algorithm}

We now present our approach for efficiently learning a patient-specific coprocessor policy. Our key insight is that we can separate the policy learning into learning the value of world actions followed by learning to produce high-value actions through stimulation.

Our approach is detailed in \cref{alg:overview} and consists of three steps: 1) training a world-action value model, $\critic$, 2) training the brain model,  $\convhat$, and 3) updating the world-action value model, $\critic$. We alternate between steps 2 and 3 during online patient interaction.

\subsection{Training world-action value model}
Our goal is to leverage a biomechanical simulator to simulate the effect of a world action $a$ on a world state $s$, given the world MDP, $M$. Via this simulation, we can learn a world policy $\pi$ for how to act optimally in the world without having to directly interact with the patient. We assume that the optimal world policy is consistent across patients (e.g., though their neural activity may differ, patients will reach the same target object by taking the same world actions) and can be readily simulated via the biomechanical simulator. We leverage this simulator to derive the Q-function, $\critic$ for the optimal  world policy, $\pi$ (\cref{alg:overview} lines 2-5).  The world policy and Q-update are defined in \cref{eq:worldpolicy} and \cref{eq:q-update} respectively.
\begin{gather}
	\pi(s) \triangleq \argmax_{a \in\A} \ex[][\big]{\Q\parens[\big]{s,a}}\label{eq:worldpolicy} \\
	\critic(s,a) \leftarrow R(s,a)+\gamma\max_{a'\in\A} \ex[][\big]{\critic\parens[\big]{s',a'}} \label{eq:q-update}
\end{gather}

In our work, we use Soft Actor-Critic (SAC) \cite{haarnoja2018soft} to learn $\critic$. However, this could be substituted for any standard actor-critic or Q-learning approach. %
We note that if a biomechanical simulator is not available for a given task, the Q-function can be equivalently learned via offline RL on a dataset of human biomechanical rollouts in the environment.
Importantly, through either a simulator or offline data, our world-action value model learns from only the biomechanical action output of a brain without relying on access to neural activity of the brain itself.

\subsection{Training Brain Model}
\label{sec:brain_model}
Given the world policy defined in \cref{eq:worldpolicy}, our objective is to next learn to transform coprocessor actions into world actions. Because of the heterogeneity and complexity of the human brain, this process cannot be easily simulated and must instead be learned online. We utilize the Q-function learned in the previous step to guide online learning (\cref{alg:overview} line 7). We aim to select stimulations $\ac$ that produce world actions of maximum value, thereby focusing learning of $\convhat$ on high-value regions of $\A$. After each patient interaction, we collect an experience, $(s, \ac, r, s')$, which we use to update $\convhat$. Our sampling strategy is defined in \cref{eq:coproc-policy}. After each interaction, we retrain $\convhat$ based upon our collected set of experiences (\cref{alg:overview} line 9).
\begin{gather}
	\pic(s) \triangleq \argmax_{\ac\in\Ac} \ex[][\big]{\Q\parens[\big]{s,\convhat(s,\ac)}}\label{eq:coproc-policy}
\end{gather}

We update the brain model via the mean-squared error loss between the predicted world action, $\hat{a}$ and the ground truth world action $a$. To effectively capture the complexity of the relationship between stimulations and world actions, we adopt a structure for $\convhat(s,\ac)$  akin to the model presented \cite{bryanNeuralCoprocessorsRestoring2023}. In \citet{bryanNeuralCoprocessorsRestoring2023}, the authors leverage a neural network to learn the mapping from stimulations to world actions from a monkey stroke dataset and show that this model is able to effectively capture the effects of stimulation on the brain

\subsection{Updating world-action value model}
The remaining issue to correct is that \cref{eq:coproc-policy} maximizes $\Q$ under the model $\convhat$. Unfortunately, $\convhat$ will not be a perfect model of the effects of stimulation and even if it were, not all actions $a\in\A$ are necessarily realizable by stimulation $\ac$. Thus, $\Q$ will be myopically over-optimistic when predicting $Q$-values from the perspective of the coprocessor agent. To solve this problem, we must continuously recalibrate the $Q$-function based on which actions can be realized by stimulation. We perform this calibration in the simulation MDP $\Mhat$ using the update in \cref{eq:q-calibrate}. This procedure is illustrated in \cref{alg:overview}, lines 10-14.
\begin{gather}
	\Q(s,a)\gets r+ \gamma \max_{\ac\in\Ac} \ex[][\big]{\Q(s',\convhat(s',\ac))} \label{eq:q-calibrate}
\end{gather}

In summary, we first train the world-action value model $\Q$ offline, and then iteratively update it while also learning $\convhat$. Repeating the last two stages (training brain model and updating world-action value model) enables us to learn an effective coprocessor policy via minimal interactions with the patient. We call our approach Coprocessor Actor Critic (\algname) and demonstrate its performance in comparison with other RL methods in the next section.

\section{Experiments}
To aid a patient in accomplishing a task such as reaching and grasping an object, a coprocessor must  learn a patient-specific stimulation policy in both an efficient and effective manner. Thus, the goal of our experimental evaluation is to 1) analyze the sample efficiency of \algname\ compared to state-of-the-art MFRL and MBRL baselines and 2) investigate the reward attained by \algname\ in comparison to these baselines. \looseness=-1

We compare \algname\ to the popular MFRL approach, Soft Actor-Critic (SAC) \cite{haarnoja2018soft},  which combines actor and critic networks with an entropy regularization term, promoting exploration in a stable and efficient manner. We choose to compare to SAC because actor-critic algorithms have been employed in prior work in learning a policy for closed-loop brain stimulation \cite{Gao,Lu}. We additionally baseline against the MBRL approach Model-Based Policy Optimization \cite{janner2019when}, which trains a model of the environment and uses both real experience and simulated experience from the model to update its policy. To assess the effectiveness of our sampling policy and the importance of updating the world-action value model, we conduct an ablation experiment and compare against \algname\ with a random  sampling policy (instead of maximizing the Q-function) as well as \algname\ without updating the world-action value model.   %
\looseness=-1

\subsection{\textit{In Silico} Evaluation Environments}
Evaluating novel RL approaches for adaptive brain stimulation \textit{in vivo} is risky for patients and may waste patients' valuable time if the approach is not successful. Thus, it is common practice to first conduct experiments \textit{in silico} (i.e., in simulation) to verify the efficacy of the approach before deploying in patients \cite{Little2013,Ashmaig2018a}. In our experiments, we investigate the ability of our method to  restore the functionality of a synthetic injured brain across a range of simulated control tasks. %
The goal in each environment is to learn a coprocessor control policy to provide the appropriate stimulation to the brain to recover function and improve performance on the tasks post-injury. Below we discuss the control tasks and the synthetic brain models employed in our \textit{in silico} experiments.

\begin{figure}
	\centering
	\includegraphics[width=\linewidth]{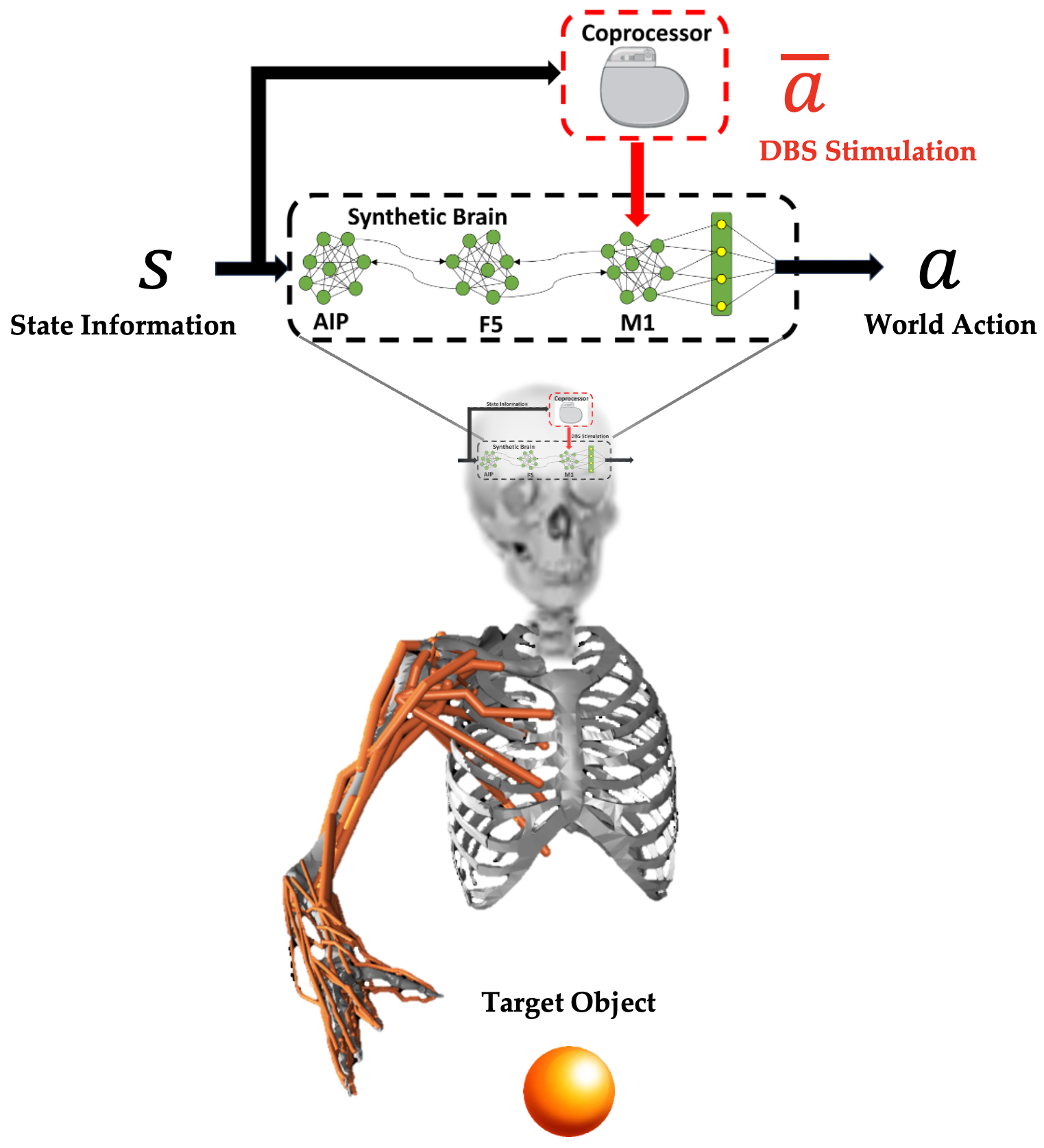}
	\caption{This figure shows the brain stimulation domain for the MyoSim Arm Reach task. We model the biomechanics of the reaching tasks using the MyoSuite physics simulator \cite{wang2022myosim}. The brain of a stroke patient is modeled via the approach described by \citet{michaelsmodel} and consists of the anterior intraparietal area (AIP), ventral premotor cortex (F5), and primary motor cortex (M1) modules. The coprocessor applies stimulation to the motor cortex (M1) which modifies the world action of the patient.  }
	\label{fig:simulation}
\end{figure}

\begin{figure*}
	\centering
	\includegraphics[width=\textwidth,trim={0 1cm 0 0}]{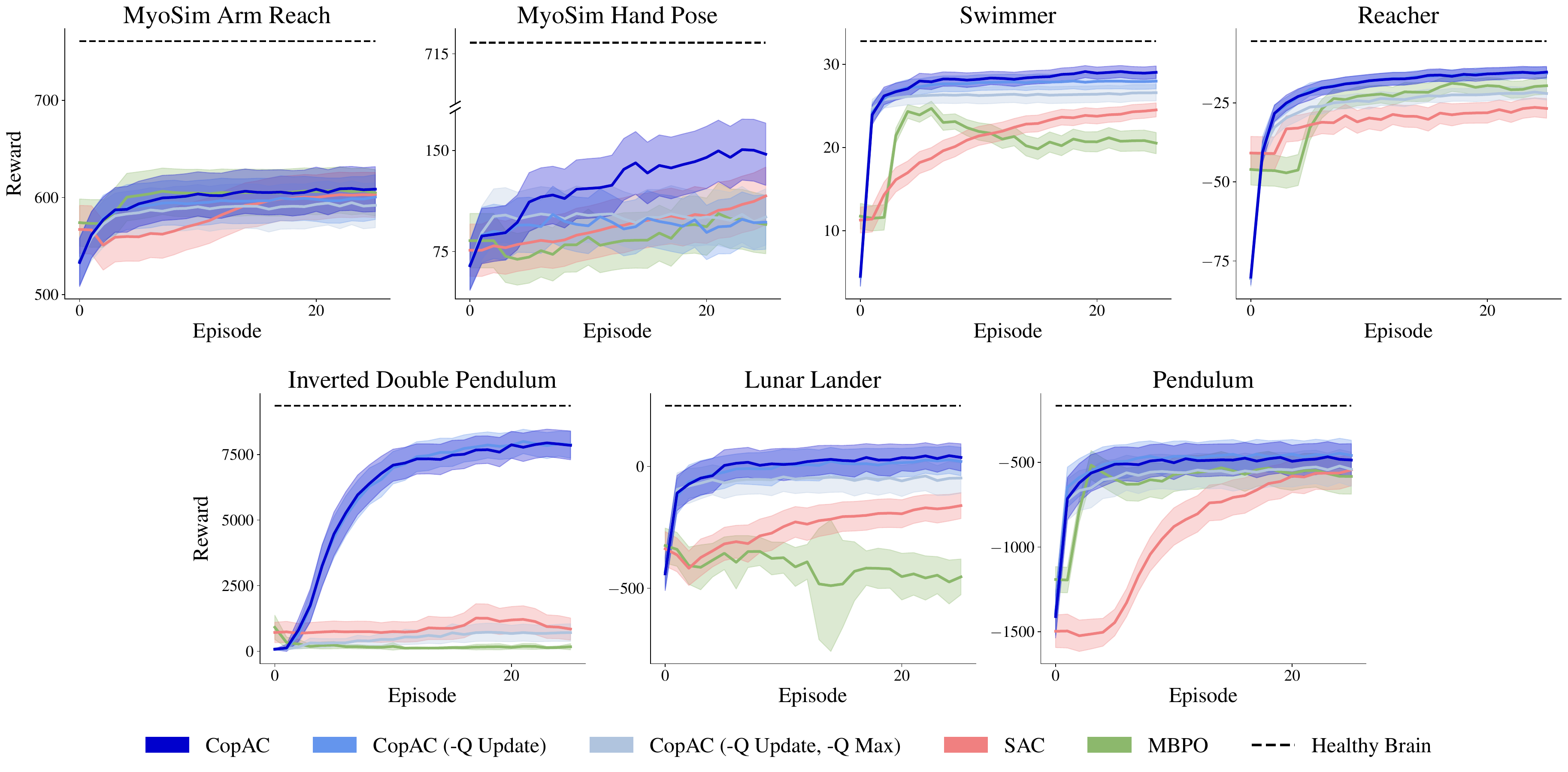}
	\caption{Evaluation results for \algname\ compared to SAC, MBPO, and ablated \algname. The dashed line represents the reward obtained by the healthy brain. }
	\label{fig:testing_results}
\end{figure*}

\textbf{Physiologically and Neurologically Realistic Stroke Domain:} Drawing on prior work in neurophysiological and biomechanical modeling \cite{michaelsmodel,wang2022myosim}, we introduce a novel simulation domain for evaluating adaptive brain stimulation policies in stroke patients (\cref{fig:simulation}). Such an \textit{in silico} evaluation requires both a neurologically realistic human brain model of a stroke patient and a high-fidelity biomechanical simulator. To simulate the biomechanics of  the human musculoskeletal system, we rely on MyoSuite \cite{wang2022myosim}, a cutting-edge simulator for biomechanical control problems based on the MuJoCo physics engine. The action space consists of individual muscle activations and the observation space consists of joint angles.

\begin{figure*}
	\centering
	\includegraphics[width=\textwidth,trim={0 1cm 0 0}]{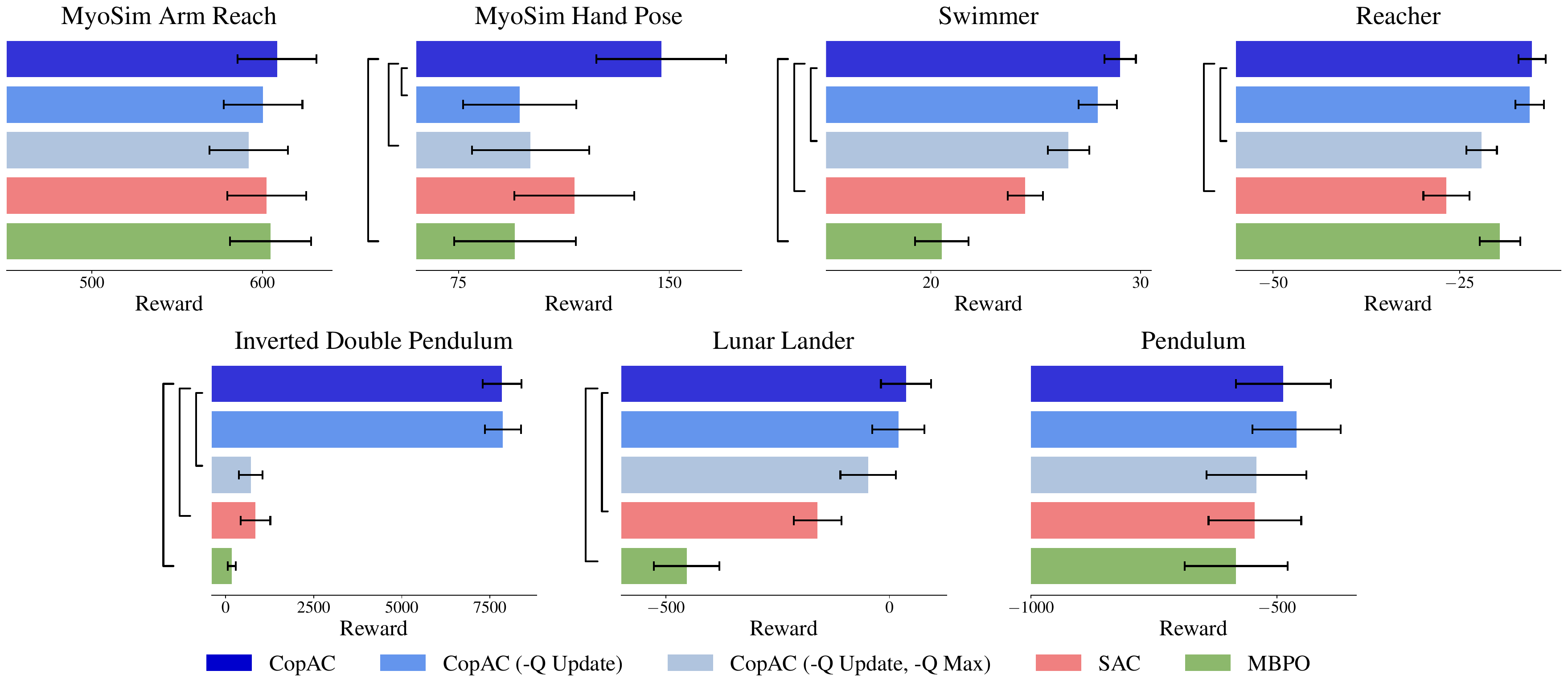}
	\caption{Evaluation results for \algname\ compared to SAC and MBPO, and ablated \algname. We display the evaluation reward after 25 episodes of training. Statistically significant differences between \algname\ and other methods are marked with brackets.}
	\label{fig:testing_results_bar}
\end{figure*}

We simulate the injured brain of a stroke patient using the cortical brain model developed by \citet{michaelsmodel}. The backbone of this model is a recurrent neural network that mimics the modular structure of the anatomical circuit encompassing the visual cortex, the anterior intraparietal area (AIP), the
ventral premotor cortex (F5), and the primary motor cortex (M1). \citet{michaelsmodel} show that the emergent neural dynamics of this model correspond to the neural responses exhibited in a non-human primate's brain. Our insight is that we can employ standard RL techniques to train this brain model to accomplish various physical tasks in the MyoSuite biomechanical simulation, thus establishing a  realistic pipeline between brain motor control signals, muscle kinematics, and biomechanical movement. \citet{michaelsmodel} additionally demonstrate that zeroing a portion of the weights in the desired brain structure can reproduce behavioral deficits caused by stroke-induced brain lesions.  %
By lesioning the simulated brain following the protocol detailed by \citet{michaelsmodel}, we induce stroke-like neurological and physiological behavior.

This method can be employed to simulate stroke patient  control strategies across various functional benchmark tasks (e.g., in-hand object manipulation, object grasping, ambulation, visual-spatial acuity, etc.). We choose to evaluate \algname\ on a reaching task that requires goal-directed functional movement (\cref{fig:simulation}). Such spatial reaching tasks, clinically known as task-related reaching training, are common benchmark tasks in which stroke patients often exhibit suboptimal control strategies \cite{THIELMAN20041613}. We also evaluate \algname\ on a spatial pose task in which the goal is to move the fingers to target locations. This task requires fine motor skills and emulates activities such as grasping an object that can be challenging for stroke patients to execute. \cite{grasp}.

Given a neurophysiologically realistic model of a stroke patient, the last step is to simulate the effects of closed-loop stimulation on the lesioned brain. \citet{bryanNeuralCoprocessorsRestoring2023} introduce a method to spatially and temporally simulate brain stimulation in the primary motor cortex to approximate the effect of \textit{in vivo} stimulation. We rely on this approach to simulate the coprocessor's effect on a stroke patient's brain.

\textbf{Standard Continuous Control Tasks:} We additionally evaluate \algname\ on a variety of standard continuous control tasks from the OpenAI Gym benchmark suite \cite{brockman2016openai}. Although the nature of these tasks is distinct from the intricate control of human movement that is typical of adaptive brain stimulation tasks, our objective in scrutinizing our approach within these domains is twofold: firstly, to showcase its adaptability in handling a variety of complex tasks with varying state and action space dimensions, and secondly, to provide a benchmark comparison against previous approaches in well-established domains.

For OpenAI Gym benchmark tasks, we simulate a control policy
generated by the brain of a stroke patient by first training
a neural network policy using standard RL techniques to
solve each Gym environment.  We then ``injure'' the policy by zeroing random weights between two hidden layers in the network. Coprocessor stimulation is applied by additively modifying the values of randomly selected neurons in the layer following lesion.

\begin{figure*}
	\centering
	\includegraphics[width=\textwidth,trim={0 1cm 0 0}]{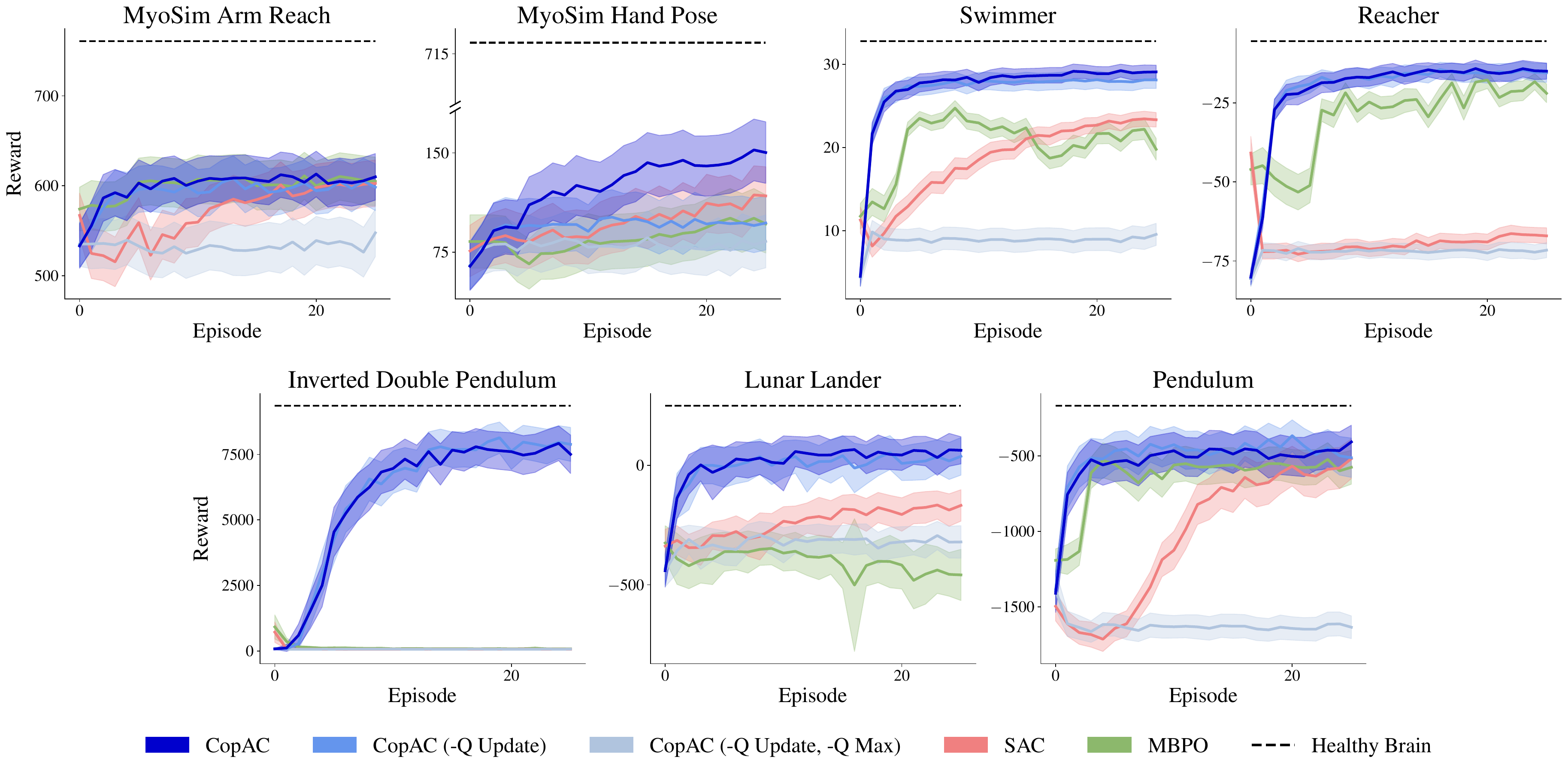}
	\caption{Training results for \algname\ compared to SAC, MBPO, and ablated \algname. The dashed line represents the reward obtained by the healthy brain. }
	\label{fig:training_results}
\end{figure*}

\subsection{Evaluation Reward}
\cref{fig:testing_results} shows a comparison of the evaluation reward after each episode of online training for each of the approaches. The dashed line represents the performance of an unaffected or ``healthy'' brain, without any lesion. We note that we do not necessarily expect a stimulation policy to be capable of achieving healthy performance, as the severity of the lesion may limit the brain's ability to reach healthy-level performance. Instead, the primary objective is to provide stimulation to assist the patient in attaining a reward as close to a healthy level as possible.

We average our results across a set of injuries ranging from 0\% to 100\% lesion of the motor cortex for 25 episodes of online interaction. We limit the number of episodes to 25 as requiring a stroke patient to perform a greater number of task repetitions would likely be too physically and mentally demanding. Despite this tight sampling constraint, \algname\ is able to quickly learn an effective strategy in less than 25 episodes in each of the control environments. On the contrary, SAC often struggles to improve performance significantly beyond random sampling within the limited time frame.  Even when SAC is able to achieve a strategy comparable to \algname, it typically requires double the number of interactions to do so. We find that MBPO  outperforms SAC  in terms of evaluation reward. In environments in which the model is simpler to learn such as MyoSim Reach and Pendulum, MBPO performs on par with \algname. However, when dealing with more complex environments such as MyoSim Pose in which each finger must be precisely manipulated, MBPO learns much slower than our approach.

We next investigate the importance of the various components of \algname\  via an ablation study. We compare the evaluation rewards of \algname\ with two modified versions: \algname\ (-Q Update) and \algname\ (-Q Update, -Q Max). \algname\ (-Q Update) does not update the world-action value model during online learning. We anticipate that not including this update will only impact performance when not all world actions are realizable due to brain lesioning. We see the most prominent benefit of updating the world-action model in the Swimmer domain. We also see an  improvement in Reacher and a small improvement towards the end of the 25 episodes in several other domains.

We next investigate the benefit of our sampling strategy  which selects the stimulation that maximizes  $\critic$, given our current knowledge of $\convhat$. To do so, we compare \algname\ to a random sampling strategy without updating the world action-value model, i.e., \algname\ (-Q update, -Q max). Because we are not focusing sampling on  high-value regions of the world actions space, we expect \algname\ (-Q update, -Q max) to learn more slowly compared to \algname. We find that a random sampling strategy under-performs in the MyoSim Pose domain compared to \algname's strategic sampling strategy and also produces lower evaluation reward in the Swimmer, Reacher,  Pendulum, and Inverted Double Pendulum domains. %

\subsection{Training Reward}
Ideally, a coprocessor should simultaneously provide stimulation that maximally assists the stroke patient given its current model of the patient while continuously refining and updating its policy. In the previous section, we established that our approach learns a superior policy with fewer samples than the baseline methods. Here we investigate if, during policy learning, we are able to achieve a higher training reward compared to baselines. Training reward is an important metric in our brain stimulation domain because it indicates how well an approach is able to aid a patient in completing the desired task during policy learning. \cref{fig:training_results} shows the training reward summed over each episode of online interaction. Due to our sampling strategy which selects stimulations that produce high-value world actions, our approach is able to achieve high task reward during learning. This means that \algname\ can successfully assist the patient in accomplishing a given task during the learning phase. In contrast, MBPO and SAC achieve a significantly lower training reward. \algname\ (-Q update, -Q max) performs poorly in all environments due to the random sampling policy.

\section{Discussion}
Our findings demonstrate the advantage of MBRL over MFRL in the domain of adaptive brain stimulation. Leveraging a model of the patient's biomechanics enables \algname\ to reduce the number of interactions and achieve more than a 10x benefit in sample efficiency in many environments. This improved sample efficiency means that patients will be able to quickly benefit from a high-quality stimulation policy. We find that the benefit of our approach is particularly apparent in domains that require complex motor movements with large action spaces (e.g., MyoSim Hand Pose) whereas domains that require more coarse movements with lower dimensional action spaces (e.g., MyoSim Arm Reach) do not produce as large of a benefit compared to baselines. This insight is important because stroke patients often struggle with complex tasks requiring fine motor skills. We show that our approach is the most effective at learning stimulation policies for these complex tasks whereas an approach like MBPO may suffice for simpler tasks that require only gross motor control. \looseness=-1

In the adaptive brain stimulation domain, high training reward is important for improving patient experience. A sampling strategy that produces off-task behavior will likely be disruptive and frustrating to a patient and could even pose a risk to the patient's safety. %
In our ablation study, we show that by leveraging the action-value model to sample in an on-policy manner, \algname\ produces higher training reward compared to random sampling. \algname's training reward contrasts with both SAC and MBRL which utilize an off-policy strategy to sample stimulations, resulting in lower  reward. Notably, \algname's strategy exhibits a significant advantage in the MyoSim Hand Pose environment. We hypothesize that this result is due to the fact that this environment has a much larger action space (39-dimensional) compared to the other environments. This result underscores the importance of a strategic sampling approach to efficiently learn a high-quality policy in environments with a substantial action space.

We find that updating the world action-value model is an important component in some domains. However, if all world actions are realizable by the injured brain or if the unrealizable actions are unimportant to task success, then updating the world action-value model may be inconsequential.

Our results show strengths of \algname\ in areas particularly crucial for adaptive brain stimulation. The learned world action-value model enables good performance throughout training, which is important for patient comfort and safety. Our approach additionally excels over baselines in complex tasks requiring fine-grained control, providing benefits in both sample efficiency and performance.

\section{Conclusion}
\textbf{Summary:} We have introduced \algname, an MBRL approach for learning a coprocessor policy for stroke patients. Our key insight is that the optimal stimulation policy is a combination of modeling optimal world actions and determining how to produce world actions via brain stimulation. Our approach leverages a simulation-derived Q-function to model the quality of world actions for a given task. We then employ this world action-value model to intelligently and efficiently learn the mapping from coprocessor stimulations to world actions. To avoid an overly-optimistic Q-function, we iteratively update the action-value model based upon our current model of the brain. We demonstrate our approach in both a novel, physiologically-motivated environment and standard control tasks. Our approach excels in terms of sample efficiency and overall task reward, surpassing both MBRL and MFRL methods across all domains. By improving sample efficiency 10-fold, we take a step towards an RL approach that can be deployed \textit{in vivo}. Our hope is to pave the way for future advancements in applying RL to closed-loop brain stimulation in real-world settings.

\textbf{Limitations:} One limitation of our work is that we only evaluate \algname\ \textit{in silico} due to the difficulty of \textit{in vivo} evaluations. The invasive nature of brain stimulation and potential negative side effects of unsafe stimulation motivate us to first validate the efficacy of our approach in simulation before deploying in the real world. We are also limited by our reliance on a realistic biomechanical simulator to enable sample efficiency. Since we test in simulation, we have access to the same environment to learn an action-value model in the first step of \algname. Our experiments therefore do not perfectly reflect real-world challenges posed by the gap in realism between simulators and human patients. Another challenge of real-world applications is the technology required (e.g., electromyography and vicon) for estimating world states and actions. Finally, although our approach minimizes patient interaction, it still requires online learning for which we have not theoretically guaranteed safety.  Such safety guarantees would be crucial for any \textit{in vivo} applications.

\textbf{Future Work:} Future work will focus on addressing these limitations. We will test how the action-value model trained in biomechanical simulation transfers to human patients in \textit{in vivo} evaluations. Additionally, we will explore offline RL for learning the action-value model in circumstances where a biomechanical simulator is not available. To improve safety and mitigate patient risk during online learning, in future work we aim to draw upon existing approaches in safe RL and we will work closely with clinical collaborators to ensure that we are safely and appropriately constraining the stimulation space  during \textit{in vivo} experiments \cite{Garca2015ACS}. %

\section*{Impact Statement} %
A major ethical concern of automated coprocessor policy learning is that adaptive brain stimulation can pose a safety risk to patients if stimulations outside of a safe region are applied to the brain. To mitigate these risks, in future work, when deploying \algname\ on real patients, we aim to work closely with clinical collaborators to ensure patient safety and appropriately constrain the action space during online learning.

Another ethical consideration involves the possibility of RL coprocessor approaches being exploited for malicious control over end-users. Unauthorized manipulation of RL policies, such as through adversarial attacks, could lead to unethical interventions and compromise the well-being of individuals with brain implants. To reduce this risk, in future work, we will draw on prior work in guarding against adversarial attacks to mitigate potential exploitation \cite{Chen2019}.

\section* {Acknowledgements}

This work was supported in part by a grant from NSF HCC, the Weill Neurohub, and by CoCoSys, one of seven centers in JUMP 2.0, a Semiconductor Research Corporation (SRC) program sponsored by DARPA.

We thank Ian Heimbuch and Matthew Bryan for early discussions on this work.

\bibliography{references}

\begin{thebibliography}{40}
\providecommand{\natexlab}[1]{#1}
\providecommand{\url}[1]{\texttt{#1}}
\expandafter\ifx\csname urlstyle\endcsname\relax
  \providecommand{\doi}[1]{doi: #1}\else
  \providecommand{\doi}{doi: \begingroup \urlstyle{rm}\Url}\fi

\bibitem[Abbeel et~al.(2006)Abbeel, Quigley, and Ng]{MBRLinaccurate}
Abbeel, P., Quigley, M., and Ng, A.~Y.
\newblock Using inaccurate models in reinforcement learning.
\newblock In \emph{23rd International Conference on Machine Learning}, ICML
  '06, pp.\  1--8. Association for Computing Machinery, 2006.
\newblock ISBN 1-59593-383-2.

\bibitem[Ashmaig et~al.(2018)Ashmaig, Connolly, Gross, and
  Mahmoudi]{Ashmaig2018a}
Ashmaig, O., Connolly, M., Gross, R.~E., and Mahmoudi, B.
\newblock {Bayesian Optimization of Asynchronous Distributed Microelectrode
  Theta Stimulation and Spatial Memory}.
\newblock \emph{Proceedings of the Annual International Conference of the IEEE
  Engineering in Medicine and Biology Society, EMBS}, 2018-July:\penalty0
  2683--2686, 2018.
\newblock ISSN 1557170X.

\bibitem[Brockman et~al.(2016)Brockman, Cheung, Pettersson, Schneider,
  Schulman, Tang, and Zaremba]{brockman2016openai}
Brockman, G., Cheung, V., Pettersson, L., Schneider, J., Schulman, J., Tang,
  J., and Zaremba, W.
\newblock Openai gym.
\newblock \emph{arXiv preprint arXiv:1606.01540}, 2016.

\bibitem[Bronte-Stewart et~al.(2020)Bronte-Stewart, Petrucci, O’Day, Afzal,
  Parker, Kehnemouyi, Wilkins, Orthlieb, and Hoffman]{Stewart}
Bronte-Stewart, H.~M., Petrucci, M.~N., O’Day, J.~J., Afzal, M.~F., Parker,
  J.~E., Kehnemouyi, Y.~M., Wilkins, K.~B., Orthlieb, G.~C., and Hoffman, S.~L.
\newblock Perspective: Evolution of control variables and policies for
  closed-loop deep brain stimulation for parkinson’s disease using
  bidirectional deep-brain-computer interfaces.
\newblock \emph{Frontiers in Human Neuroscience}, 14, August 2020.
\newblock ISSN 16625161.

\bibitem[Bryan et~al.(2023)Bryan, Jiang, and
  Rao]{bryanNeuralCoprocessorsRestoring2023}
Bryan, M.~J., Jiang, L.~P., and Rao, R. P.~N.
\newblock Neural co-processors for restoring brain function: Results from a
  cortical model of grasping.
\newblock \emph{Journal of Neural Engineering}, 20\penalty0 (3), May 2023.
\newblock ISSN 1741-2552.

\bibitem[Buhmann et~al.(2017)Buhmann, Huckhagel, Engel, Gulberti, Hidding,
  Poetter-Nerger, Goerendt, Ludewig, Braass, Choe, Krajewski, Oehlwein,
  Mittmann, Engel, Gerloff, Westphal, Köppen, Moll, and Hamel]{grasp}
Buhmann, C., Huckhagel, T., Engel, K., Gulberti, A., Hidding, U.,
  Poetter-Nerger, M., Goerendt, I., Ludewig, P., Braass, H., Choe, C.~U.,
  Krajewski, K., Oehlwein, C., Mittmann, K., Engel, A.~K., Gerloff, C.,
  Westphal, M., Köppen, J.~A., Moll, C. K.~E., and Hamel, W.
\newblock Adverse events in deep brain stimulation: A retrospective long-term
  analysis of neurological, psychiatric and other occurrences.
\newblock \emph{PLoS ONE}, 12, July 2017.
\newblock ISSN 19326203.

\bibitem[Chen et~al.(2019)Chen, Liu, Xiang, Niu, Tong, and Han]{Chen2019}
Chen, T., Liu, J., Xiang, Y., Niu, W., Tong, E., and Han, Z.
\newblock Adversarial attack and defense in reinforcement learning-from {AI}
  security view.
\newblock \emph{Cybersecurity}, 2, December 2019.
\newblock ISSN 25233246.

\bibitem[Choi et~al.(2023)Choi, Park, Rha, Nam, Jo, and Kim]{Choi2023}
Choi, H., Park, D., Rha, D.~W., Nam, H.~S., Jo, Y.~J., and Kim, D.~Y.
\newblock Kinematic analysis of movement patterns during a reach-and-grasp task
  in stroke patients.
\newblock \emph{Frontiers in Neurology}, 14, 2023.
\newblock ISSN 16642295.

\bibitem[Coventry \& Bartlett(2023)Coventry and Bartlett]{Coventry}
Coventry, B.~S. and Bartlett, E.~L.
\newblock Closed-loop reinforcement learning based deep brain stimulation using
  {SpikerNet}: A computational model.
\newblock In \emph{2023 11th International {IEEE/EMBS} Conference on Neural
  Engineering {(NER)}}, pp.\  1--4, 2023.

\bibitem[Cunha et~al.(2015)Cunha, Boschen, {Gómez-A}, Ross, Gibson, Min, Lee,
  and Blaha]{dacunhaSophisticatedBasalGanglia2015}
Cunha, C.~D., Boschen, S.~L., {Gómez-A}, A., Ross, E.~K., Gibson, W. S.~J.,
  Min, H.-K., Lee, K.~H., and Blaha, C.~D.
\newblock Toward sophisticated basal ganglia neuromodulation: {{Review}} on
  basal ganglia deep brain stimulation.
\newblock \emph{Neuroscience \& Biobehavioral Reviews}, 58:\penalty0 186--210,
  November 2015.
\newblock ISSN 0149-7634.

\bibitem[Delp et~al.(2007)Delp, Anderson, Arnold, Loan, Habib, John,
  Guendelman, and Thelen]{opensim}
Delp, S.~L., Anderson, F.~C., Arnold, A.~S., Loan, P., Habib, A., John, C.~T.,
  Guendelman, E., and Thelen, D.~G.
\newblock {OpenSim}: Open-source software to create and analyze dynamic
  simulations of movement.
\newblock \emph{IEEE Transactions on Biomedical Engineering}, 54\penalty0
  (11):\penalty0 1940--1950, 2007.

\bibitem[Dulac-Arnold et~al.(2021)Dulac-Arnold, Levine, Mankowitz, Li,
  Paduraru, Gowal, and Hester]{arnold}
Dulac-Arnold, G., Levine, N., Mankowitz, D.~J., Li, J., Paduraru, C., Gowal,
  S., and Hester, T.
\newblock Challenges of real-world reinforcement learning: definitions,
  benchmarks and analysis.
\newblock \emph{Machine Learning}, 110:\penalty0 2419--2468, September 2021.
\newblock ISSN 15730565.

\bibitem[Elias et~al.(2018)Elias, Namasivayam, and Lozano]{ELIAS20183}
Elias, G. J.~B., Namasivayam, A.~A., and Lozano, A.~M.
\newblock Deep brain stimulation for stroke: Current uses and future
  directions.
\newblock \emph{Brain Stimulation}, 11\penalty0 (1):\penalty0 3--28, 2018.
\newblock ISSN 1935-861X.

\bibitem[Frey et~al.(2022)Frey, Cagle, Johnson, Wong, Hilliard, Butson, Okun,
  and de~Hemptinne]{Frey2022}
Frey, J., Cagle, J., Johnson, K.~A., Wong, J.~K., Hilliard, J.~D., Butson,
  C.~R., Okun, M.~S., and de~Hemptinne, C.
\newblock Past, present, and future of deep brain stimulation: Hardware,
  software, imaging, physiology and novel approaches.
\newblock \emph{Frontiers in Neurology}, 13, March 2022.
\newblock ISSN 16642295.

\bibitem[Ganguly et~al.(2022)Ganguly, Khanna, Morecraft, and
  Lin]{gangulyModulationNeuralCofiring2022}
Ganguly, K., Khanna, P., Morecraft, R.~J., and Lin, D.~J.
\newblock Modulation of neural co-firing to enhance network transmission and
  improve motor function after stroke.
\newblock \emph{Neuron}, 110\penalty0 (15):\penalty0 2363--2385, August 2022.
\newblock ISSN 08966273.

\bibitem[Gao et~al.(2023)Gao, Schmidt, Chowdhury, Feng, Peters, Genty, Grill,
  Turner, and Pajic]{Gao}
Gao, Q., Schmidt, S.~L., Chowdhury, A., Feng, G., Peters, J.~J., Genty, K.,
  Grill, W.~M., Turner, D.~A., and Pajic, M.
\newblock Offline learning of closed-loop deep brain stimulation controllers
  for parkinson disease treatment.
\newblock In \emph{{ACM/IEEE 14th} International Conference on Cyber-Physical
  Systems (with {CPS-IoT} Week 2023)}, ICCPS '23, pp.\  44--55. Association for
  Computing Machinery, 2023.
\newblock ISBN 979-8400-70-0-3-6-1.

\bibitem[García \& Fernández(2015)García and Fernández]{Garca2015ACS}
García, J. and Fernández, F.
\newblock A comprehensive survey on safe reinforcement learning.
\newblock \emph{J. Mach. Learn. Res.}, 16:\penalty0 1437--1480, 2015.

\bibitem[Ghasemi et~al.(2018)Ghasemi, Sahraee, and Mohammadi]{Ghasemi2018}
Ghasemi, P., Sahraee, T., and Mohammadi, A.
\newblock Closed-and open-loop deep brain stimulation: Methods, challenges,
  current and future aspects.
\newblock \emph{Journal of Biomedical Physics and Engineering}, 8:\penalty0
  209--216, 2018.

\bibitem[Groiss et~al.(2009)Groiss, Wojtecki, Sudmeyer, and
  Schnitzler]{Groiss2009}
Groiss, S.~J., Wojtecki, L., Sudmeyer, M., and Schnitzler, A.
\newblock Deep brain stimulation in parkinson-s disease.
\newblock \emph{Therapeutic Advances in Neurological Disorders}, 2:\penalty0
  379--391, 2009.
\newblock ISSN 17562856.

\bibitem[Haarnoja et~al.(2018)Haarnoja, Zhou, Abbeel, and
  Levine]{haarnoja2018soft}
Haarnoja, T., Zhou, A., Abbeel, P., and Levine, S.
\newblock Soft actor-critic: Off-policy maximum entropy deep reinforcement
  learning with a stochastic actor.
\newblock In \emph{International {{Conference}} on {{Machine Learning}}}, 2018.

\bibitem[Hatem et~al.(2016)Hatem, Saussez, della Faille, Prist, Zhang, Dispa,
  and Bleyenheuft]{Hatem2016}
Hatem, S.~M., Saussez, G., della Faille, M., Prist, V., Zhang, X., Dispa, D.,
  and Bleyenheuft, Y.
\newblock Rehabilitation of motor function after stroke: A multiple systematic
  review focused on techniques to stimulate upper extremity recovery.
\newblock \emph{Frontiers in Human Neuroscience}, 10, September 2016.
\newblock ISSN 16625161.

\bibitem[Hu \& Stead(2014)Hu and Stead]{hu2014deep}
Hu, W. and Stead, M.
\newblock Deep brain stimulation for dystonia.
\newblock \emph{Translational Neurodegeneration}, 3:\penalty0 2, January 2014.
\newblock ISSN 2047-9158.
\newblock \doi{10.1186/2047-9158-3-2}.

\bibitem[Huang(2020)]{MFRL}
Huang, Q.
\newblock Model-based or model-free, a review of approaches in reinforcement
  learning.
\newblock In \emph{2020 International Conference on Computing and Data Science
  {(CDS)}}, pp.\  219--221, 2020.

\bibitem[Ingwersen et~al.(2021)Ingwersen, Wolf, Birke, Schlemm, Bartling,
  Bender, Meyer, Nolte, Ottes, Pade, Peller, Steinmetz, Gerloff, and
  Thomalla]{Ingwersen2021}
Ingwersen, T., Wolf, S., Birke, G., Schlemm, E., Bartling, C., Bender, G.,
  Meyer, A., Nolte, A., Ottes, K., Pade, O., Peller, M., Steinmetz, J.,
  Gerloff, C., and Thomalla, G.
\newblock Long-term recovery of upper limb motor function and self-reported
  health: results from a multicenter observational study 1 year after
  discharge from rehabilitation.
\newblock \emph{Neurological Research and Practice}, 3, December 2021.
\newblock ISSN 25243489.

\bibitem[Janner et~al.(2019)Janner, Fu, Zhang, and Levine]{janner2019when}
Janner, M., Fu, J., Zhang, M., and Levine, S.
\newblock When to {{Trust Your Model}}: {{Model-Based Policy Optimization}}.
\newblock In \emph{Advances in {{Neural Information Processing Systems}}},
  volume~32. Curran Associates, Inc., 2019.

\bibitem[Khanna et~al.(2021)Khanna, Totten, Novik, Roberts, Morecraft, and
  Ganguly]{khannaLowfrequencyStimulationEnhances2021}
Khanna, P., Totten, D., Novik, L., Roberts, J., Morecraft, R.~J., and Ganguly,
  K.
\newblock Low-frequency stimulation enhances ensemble co-firing and dexterity
  after stroke.
\newblock \emph{Cell}, 184\penalty0 (4):\penalty0 912--930.e20, February 2021.
\newblock ISSN 00928674.

\bibitem[Little et~al.(2013{\natexlab{a}})Little, Pogosyan, Neal, Zavala,
  Zrinzo, Hariz, Foltynie, Limousin, Ashkan, FitzGerald, Green, Aziz, and
  Brown]{Little2013}
Little, S., Pogosyan, A., Neal, S., Zavala, B., Zrinzo, L., Hariz, M.,
  Foltynie, T., Limousin, P., Ashkan, K., FitzGerald, J., Green, A.~L., Aziz,
  T.~Z., and Brown, P.
\newblock Adaptive deep brain stimulation in advanced parkinson disease.
\newblock \emph{Annals of neurology}, 74:\penalty0 449--457,
  2013{\natexlab{a}}.
\newblock ISSN 15318249.

\bibitem[Little et~al.(2013{\natexlab{b}})Little, Pogosyan, Neal, Zavala,
  Zrinzo, Hariz, Foltynie, Limousin, Ashkan, FitzGerald,
  et~al.]{littleAdaptiveDeepBrain2013}
Little, S., Pogosyan, A., Neal, S., Zavala, B., Zrinzo, L., Hariz, M.,
  Foltynie, T., Limousin, P., Ashkan, K., FitzGerald, J., et~al.
\newblock Adaptive deep brain stimulation in advanced {{Parkinson}} disease.
\newblock \emph{Annals of Neurology}, 74\penalty0 (3):\penalty0 449--457,
  September 2013{\natexlab{b}}.
\newblock ISSN 1531-8249.

\bibitem[Lozano et~al.(2019)Lozano, Lipsman, Bergman, Brown, Chabardes, Chang,
  Matthews, McIntyre, Schlaepfer, Schulder, Temel, Volkmann, and
  Krauss]{Lozano2019}
Lozano, A.~M., Lipsman, N., Bergman, H., Brown, P., Chabardes, S., Chang,
  J.~W., Matthews, K., McIntyre, C.~C., Schlaepfer, T.~E., Schulder, M., Temel,
  Y., Volkmann, J., and Krauss, J.~K.
\newblock Deep brain stimulation: current challenges and future directions.
\newblock \emph{Nature Reviews Neurology}, 15:\penalty0 148--160, March 2019.
\newblock ISSN 17594766.

\bibitem[Lu et~al.(2020{\natexlab{a}})Lu, Wei, Che, Wang, and Loparo]{Lu}
Lu, M., Wei, X., Che, Y., Wang, J., and Loparo, K.~A.
\newblock Application of reinforcement learning to deep brain stimulation in a
  computational model of parkinson’s disease.
\newblock \emph{IEEE Transactions on Neural Systems and Rehabilitation
  Engineering}, 28\penalty0 (1):\penalty0 339--349, 2020{\natexlab{a}}.

\bibitem[Lu et~al.(2020{\natexlab{b}})Lu, Wei, Che, Wang, and
  Loparo]{luApplicationReinforcementLearning2020a}
Lu, M., Wei, X., Che, Y., Wang, J., and Loparo, K.~A.
\newblock Application of {{Reinforcement Learning}} to {{Deep Brain
  Stimulation}} in a {{Computational Model}} of {{Parkinson}}'s {{Disease}}.
\newblock \emph{IEEE transactions on neural systems and rehabilitation
  engineering: a publication of the IEEE Engineering in Medicine and Biology
  Society}, 28\penalty0 (1):\penalty0 339--349, January 2020{\natexlab{b}}.
\newblock ISSN 1558-0210.

\bibitem[Michaels et~al.(2020)Michaels, Schaffelhofer, Agudelo-Toro, and
  Scherberger]{michaelsmodel}
Michaels, J.~A., Schaffelhofer, S., Agudelo-Toro, A., and Scherberger, H.
\newblock A goal-driven modular neural network predicts parietofrontal neural
  dynamics during grasping.
\newblock \emph{The Proceedings of the National Academy of Sciences}, 2020.

\bibitem[Moerland et~al.(2023)Moerland, Broekens, Plaat, and
  Jonker]{moerland2023modelbased}
Moerland, T.~M., Broekens, J., Plaat, A., and Jonker, C.~M.
\newblock Model-based reinforcement learning: A survey.
\newblock In \emph{Found. {{Trends Mach}}. {{Learn}}.} arXiv, 2023.
\newblock \doi{10.1561/2200000086}.

\bibitem[Oehrn et~al.(2023)Oehrn, Cernera, Hammer, Shcherbakova, Yao, Hahn,
  Wang, Ostrem, Little, and Starr]{Oehrn2023}
Oehrn, C.~R., Cernera, S., Hammer, L.~H., Shcherbakova, M., Yao, J., Hahn, A.,
  Wang, S., Ostrem, J.~L., Little, S., and Starr, P.~A.
\newblock Personalized chronic adaptive deep brain stimulation outperforms
  conventional stimulation in parkinson's disease.
\newblock \emph{medRxiv : the preprint server for health sciences}, August
  2023.

\bibitem[Rao(2019)]{Rao2019}
Rao, R.~P.
\newblock Towards neural co-processors for the brain: combining decoding and
  encoding in brain–computer interfaces.
\newblock \emph{Current Opinion in Neurobiology}, 55:\penalty0 142--151, April
  2019.
\newblock ISSN 18736882.

\bibitem[Schrum et~al.(2022)Schrum, Connolly, Cole, Ghetiya, Gross, and
  Gombolay]{SafeMetal}
Schrum, M., Connolly, M.~J., Cole, E., Ghetiya, M., Gross, R., and Gombolay,
  M.~C.
\newblock Meta-active learning in probabilistically safe optimization.
\newblock \emph{IEEE Robotics and Automation Letters}, 7\penalty0 (4):\penalty0
  10713--10720, 2022.

\bibitem[Thielman et~al.(2004)Thielman, Dean, and Gentile]{THIELMAN20041613}
Thielman, G.~T., Dean, C.~M., and Gentile, A.~M.
\newblock Rehabilitation of reaching after stroke: Task-related training versus
  progressive resistive.
\newblock \emph{Archives of Physical Medicine and Rehabilitation}, 85\penalty0
  (10):\penalty0 1613--1618, 2004.
\newblock ISSN 0003-9993.

\bibitem[Valencia et~al.(2023)Valencia, Jia, Li, Hayashi, Lecchi, Terezakis,
  Gee, Liarokapis, MacDonald, and Williams]{valencia}
Valencia, D., Jia, J., Li, R., Hayashi, A., Lecchi, M., Terezakis, R., Gee, T.,
  Liarokapis, M., MacDonald, B.~A., and Williams, H.
\newblock Comparison of model-based and model-free reinforcement learning for
  real-world dexterous robotic manipulation tasks.
\newblock In \emph{2023 {IEEE} International Conference on Robotics and
  Automation {(ICRA)}}, pp.\  871--878, 2023.

\bibitem[Visanji et~al.(2022)Visanji, Ghani, Yu, Kakhki, Sato, Moreno,
  Naranian, Poon, Abdollahi, Naghibzadeh, Rajalingam, Lozano, Kalia, Hodaie,
  Cohn, Statucka, Boutet, Elias, Germann, Munhoz, Lang, Gan-Or, Rogaeva, and
  Fasano]{Visanji2022}
Visanji, N.~P., Ghani, M., Yu, E., Kakhki, E.~G., Sato, C., Moreno, D.,
  Naranian, T., Poon, Y.~Y., Abdollahi, M., Naghibzadeh, M., Rajalingam, R.,
  Lozano, A.~M., Kalia, S.~K., Hodaie, M., Cohn, M., Statucka, M., Boutet, A.,
  Elias, G. J.~B., Germann, J., Munhoz, R., Lang, A.~E., Gan-Or, Z., Rogaeva,
  E., and Fasano, A.
\newblock Axial impairment following deep brain stimulation in parkinson's
  disease: A surgicogenomic approach.
\newblock \emph{Journal of Parkinson's Disease}, 12:\penalty0 117--128, 2022.
\newblock ISSN 1877718X.

\bibitem[Wang et~al.(2022)Wang, Caggiano, Durandau, Sartori, and
  Kumar]{wang2022myosim}
Wang, H., Caggiano, V., Durandau, G., Sartori, M., and Kumar, V.
\newblock {{MyoSim}}: {{Fast}} and physiologically realistic {{MuJoCo}} models
  for musculoskeletal and exoskeletal studies.
\newblock In \emph{2022 {{International Conference}} on {{Robotics}} and
  {{Automation}} ({{ICRA}})}, pp.\  8104--8111, May 2022.
\newblock \doi{10.1109/ICRA46639.2022.9811684}.

\end{thebibliography}
\bibliographystyle{icml2024}

\newpage
\appendix
\onecolumn

\section*{Appendix}

\section{Learning $\critic$ with offline RL}

Our method hinges on learning $\critic$ without requiring online patient interactions. While we demonstrate that we can leverage a biomechanical simulator to learn $\critic$, in some instances we may not have access to a high-fidelity simulator.  In these cases, we consider the use of an existing dataset to train $\critic$ through offline RL. This approach assumes that we have access to a historical coprocessor dataset from stimulation policies previously deployed on the patient. We use Conservative Soft Actor Critic to learn $\critic$ from this dataset. Once $\critic$ is learned from the offline data, we follow the same procedure for learning $\convhat$ as discussed in Section \ref{sec:brain_model}.

\begin{figure}[H]
	\centering
	\includegraphics[width=\textwidth,trim={0 0.5cm 0 1cm}]{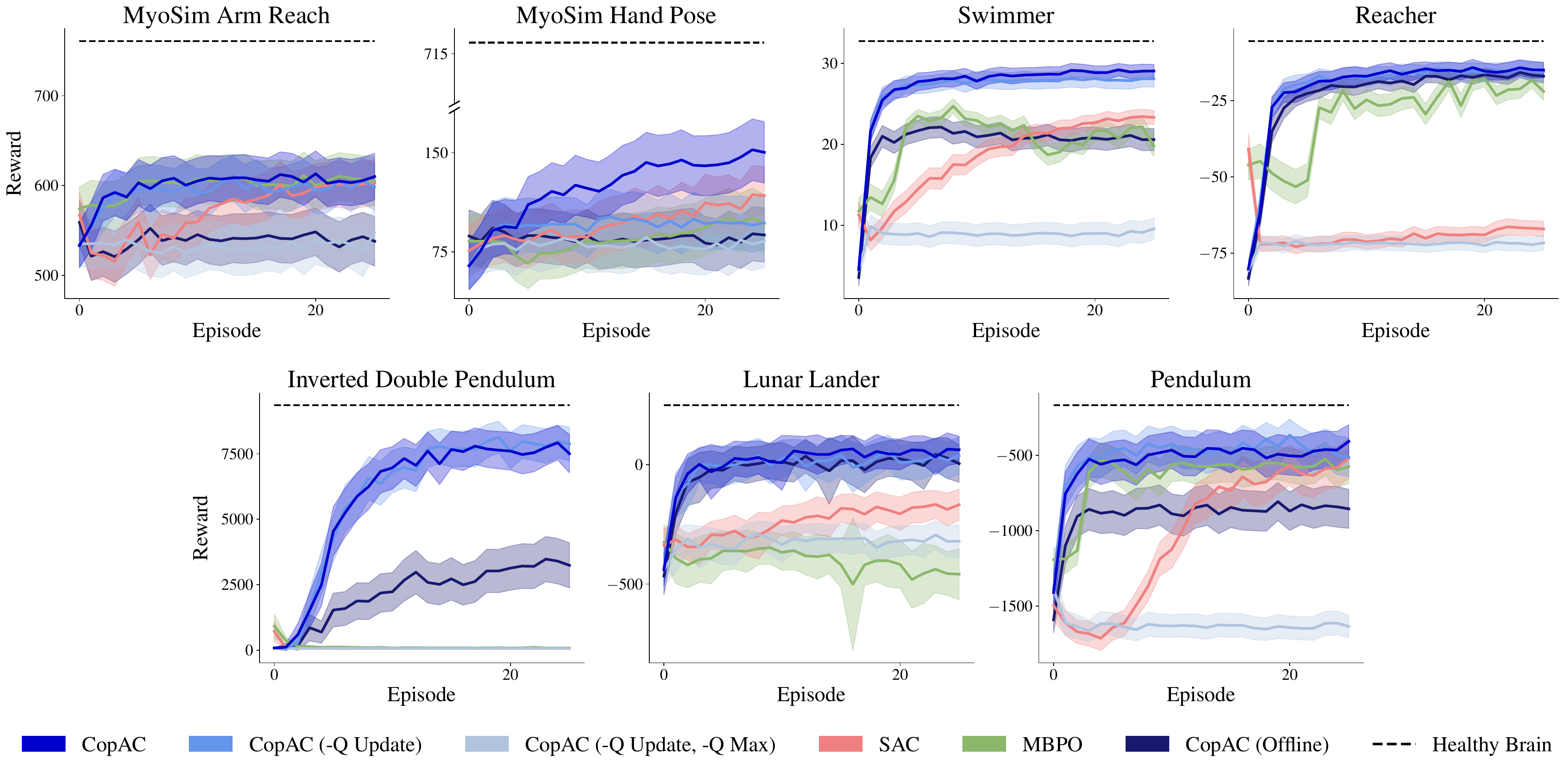}
	\caption{Training reward for \algname\ compared to SAC, MBPO, and ablated \algname. The dashed line represents the reward obtained by the healthy brain.}
	\label{fig:offline_results}
\end{figure}

\cref{fig:offline_results} shows the training reward for our approach when $\critic$ is learned via offline RL compared to baselines. We show that CopAC (offline) performs better than the baselines in most environments but performs slightly worse than CopAC  when $\critic$ is learned via simulation. This outcome supports the viability of offline RL as an alternative approach. However, it suggests that using a biomechanical simulator, when available, is likely a better option for learning $\critic$.  In future work we aim to investigate how the amount of data and the suboptimality of the policy used to collect the data affects performance.

\section{$\convhat$ training details}

We train $\convhat$ for 75 epochs with a learning rate of $5\cdot 10^{-3}$. The architecture consists of three hidden layers with ReLU activations consisting of 64, 32, and 8 neurons.

\section{Robustness to initialization of $\convhat$}

\begin{figure}[H]
	\centering
	\includegraphics[width=\textwidth,trim={0 1cm 0 0}]{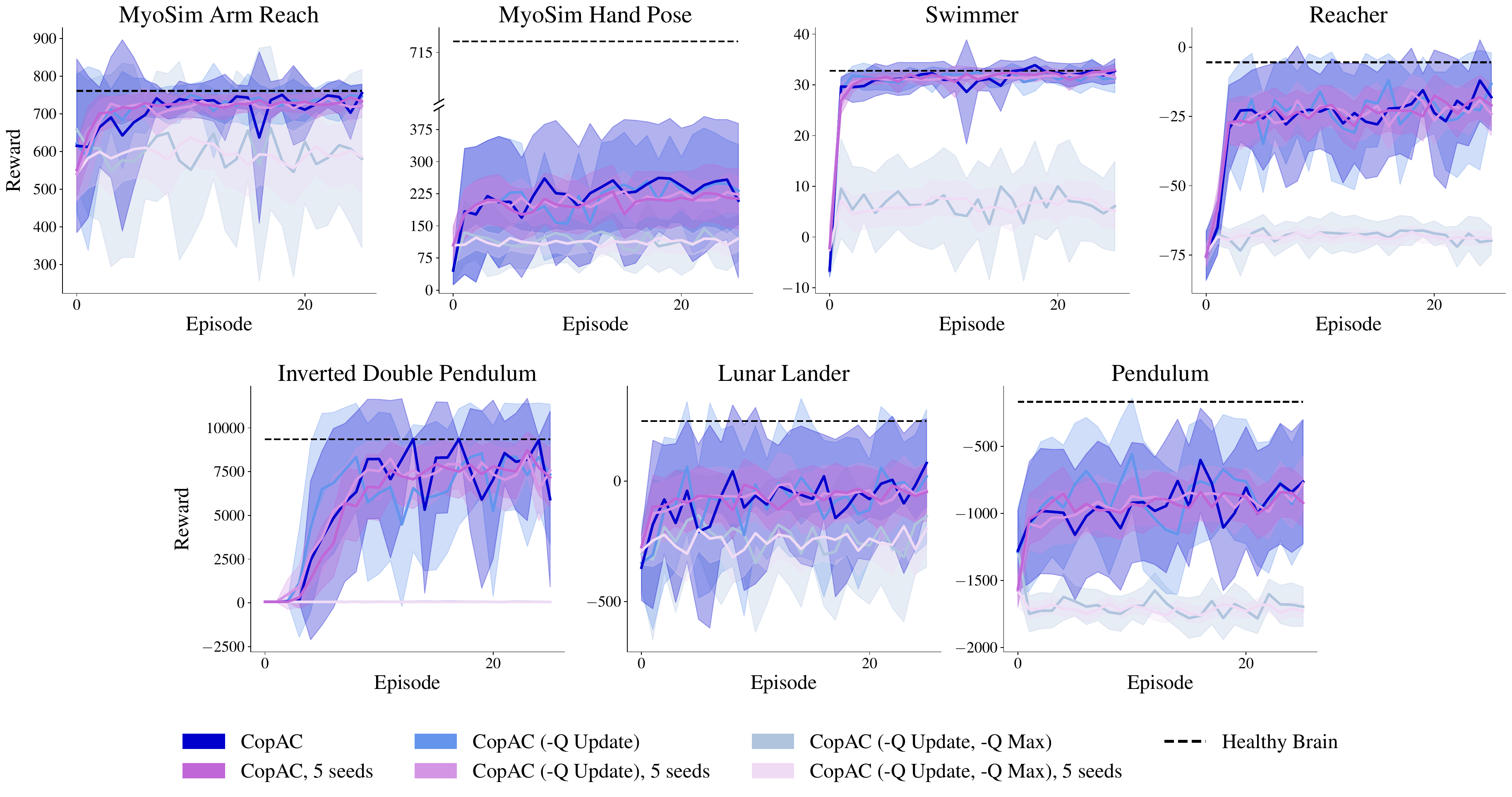}
	\caption{Training reward for \algname\ and ablations. Results from a single seed are displayed alongside results averaged across random seeds.}
	\label{fig:random_init_training_results}
\end{figure}

\begin{figure}[H]
	\centering
	\includegraphics[width=\textwidth,trim={0 1cm 0 0}]{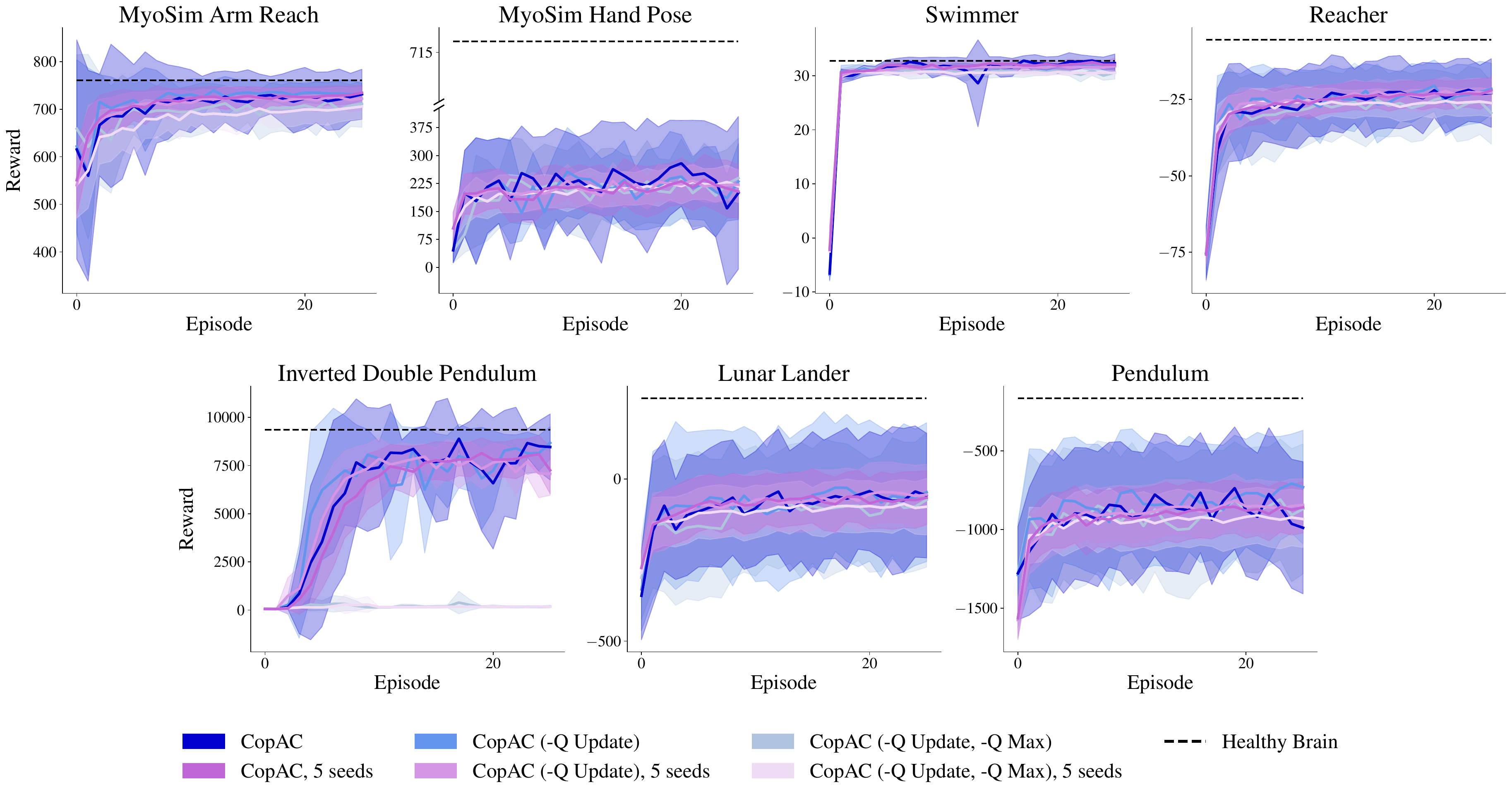}
	\caption{Evaluation reward for \algname\ and ablations.
	Results from a single seed are displayed alongside results averaged across random seeds.}
	\label{fig:random_init_testing_results}
\end{figure}

To validate that \algname\ is robust to the initialization of $\convhat$, we additionally run \algname\ and ablations across 5 random seeds. Results are displayed in \cref{fig:random_init_training_results,fig:random_init_testing_results}. Here, we only use a single brain for each environment rather than taking the average across multiple brains as in our other experiments.

\section{Comparison with inverse brain model coprocessor}

We additionally compare \algname\ to a baseline approach using an inverse brain model to select stimulations. We learn the inverse brain model $\invconvhat: S \times \A \to \Ac$ that maps world actions to the stimulations that would have induced them. The inverse brain model coprocessor policy selects stimulations with $\invconvhat$ to induce world actions returned by the optimal world policy $\pi$, as defined in \cref{eq:inverse-coproc-policy}.

\begin{gather}
	\piinv(s) \triangleq \invconvhat(s, \pi(s))\label{eq:inverse-coproc-policy}
\end{gather}

\floatname{algorithm}{\color{editcolor}Algorithm}
\begin{algorithm}[H]
	\color{editcolor}\begin{algorithmic}[1]
		\caption{Inverse Brain Model Coprocessor} \label{alg:f_inverse}
		\REQUIRE world MDP $M=(\S,\A,P,R)$, $s_0\sim p_0$
		\REQUIRE stimulation space $\Ac$, injured brain $\conv$
		\STATE initialize $\invconvhat$

		\WHILE {access to simulator}
		\STATE rollout $\pi$ in $M$ with experiences $(s, a, r, s')$
		\STATE fit $\critic(s,a)$ with \cref{eq:q-update}
		\ENDWHILE

		\WHILE {access to injured brain}
		\STATE rollout $\piinv$ in $\Mbar$ with experiences $(s, \ac, r, s')$
		\STATE $a\gets\conv(s,\ac)$
		\STATE fit $\invconvhat(s,a)$ to $\ac$
		\ENDWHILE
		\STATE \textbf{return} $\piinv$
	\end{algorithmic}
\end{algorithm}

\begin{figure}[H]
	\centering
	\includegraphics[width=\textwidth,trim={0 1cm 0 0}]{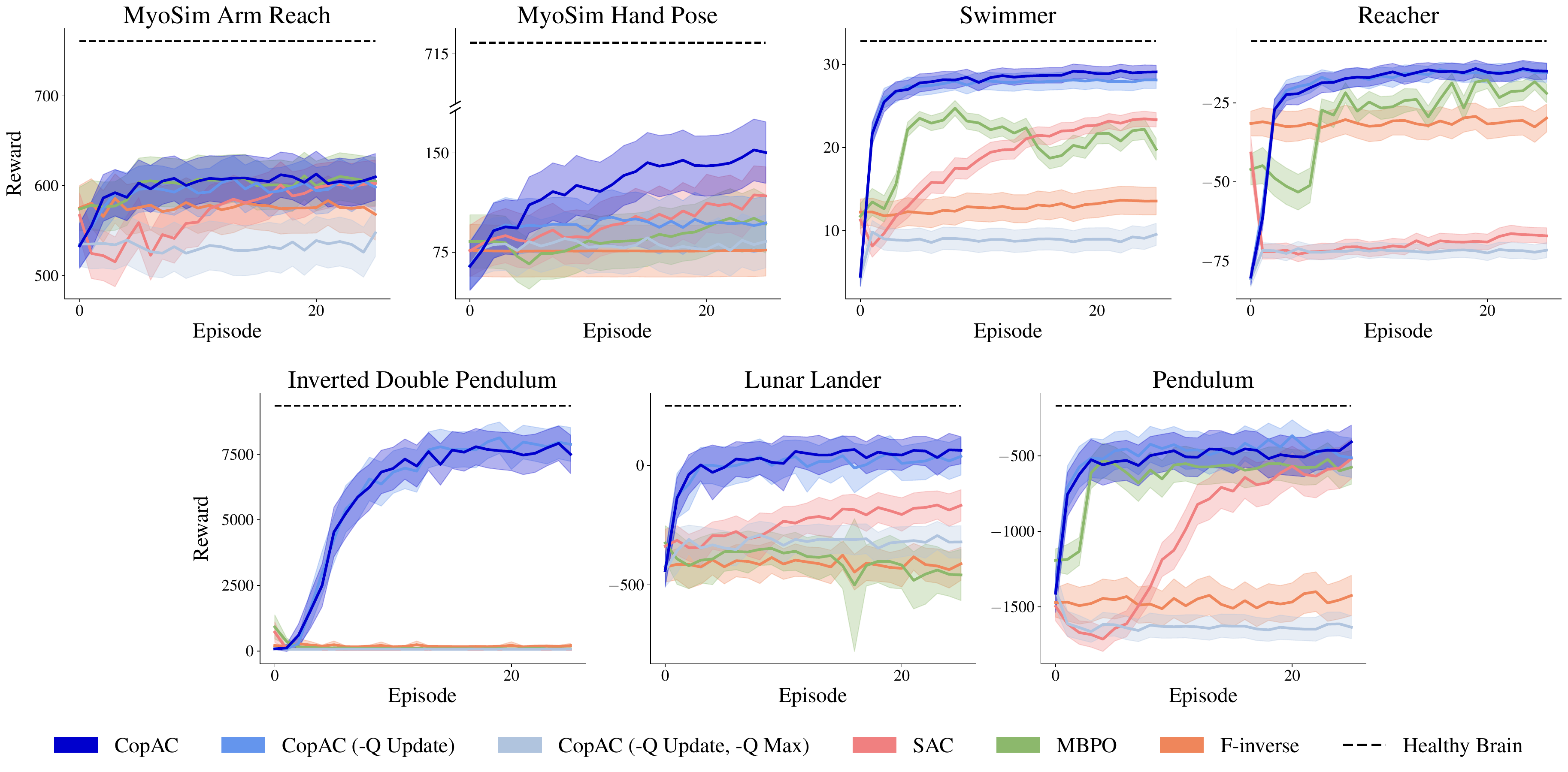}
	\caption{Training results for inverse brain model coprocessor compared to \algname, ablated \algname, and baselines.}
	\label{fig:f_inverse_training_results}
\end{figure}

\begin{figure}[H]
	\centering
	\includegraphics[width=\textwidth,trim={0 1cm 0 0}]{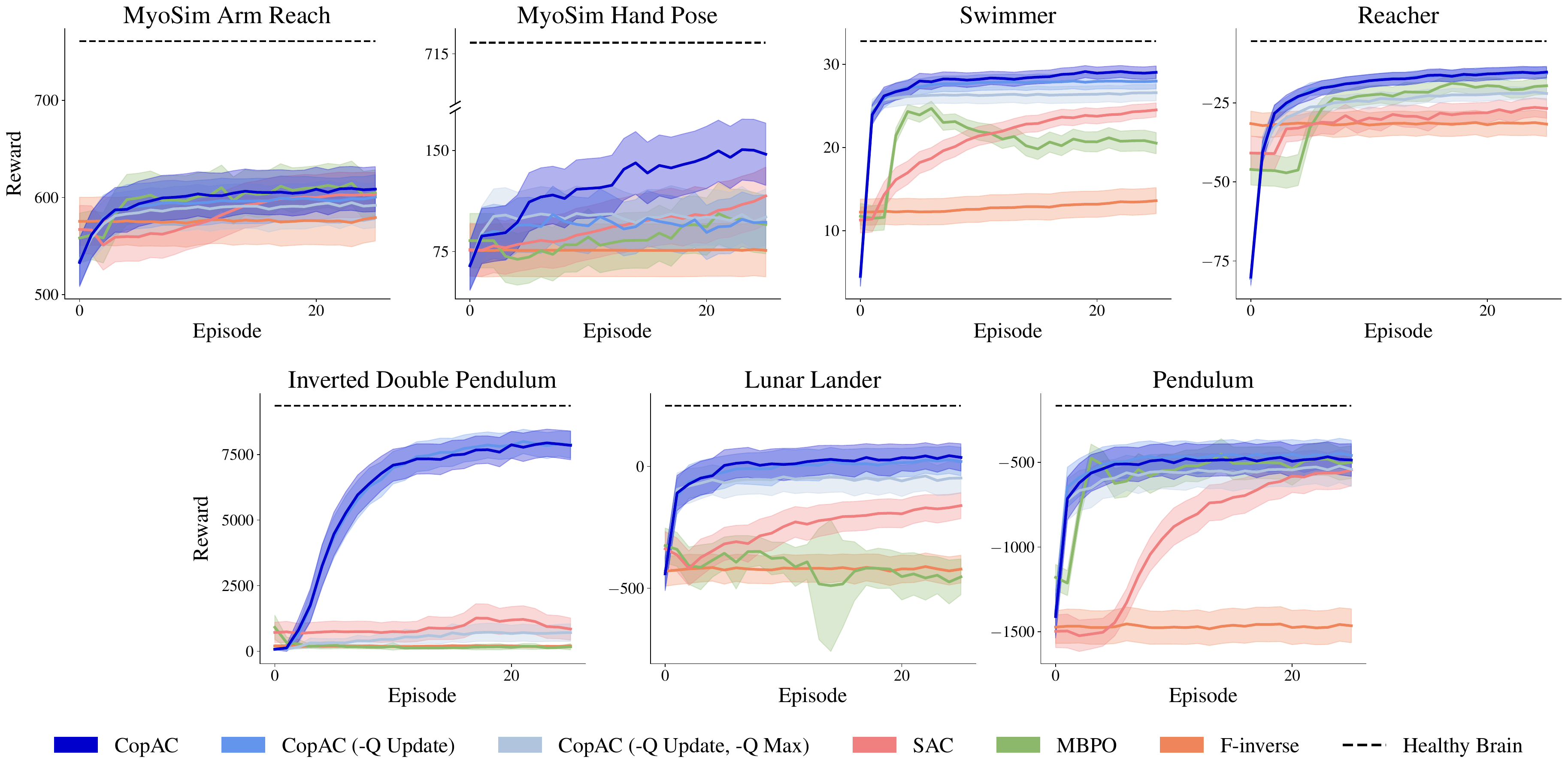}
	\caption{Evaluation results for inverse brain model coprocessor compared to \algname, ablated \algname, and baselines.}
	\label{fig:f_inverse_testing_results}
\end{figure}

The inverse brain model approach is presented in \cref{alg:f_inverse}. We evaluate it against \algname\ and show a comparison of their performance during training and evaluation in \cref{fig:f_inverse_training_results,fig:f_inverse_testing_results}. We find that CopAC is able to achieve a higher reward and better sample efficiency compared to the inverse brain model.

\section{Robustness to sim to real gap}
\begin{figure}[H]
	\centering
	\includegraphics[width=10cm,trim={0 1cm 0 0}]{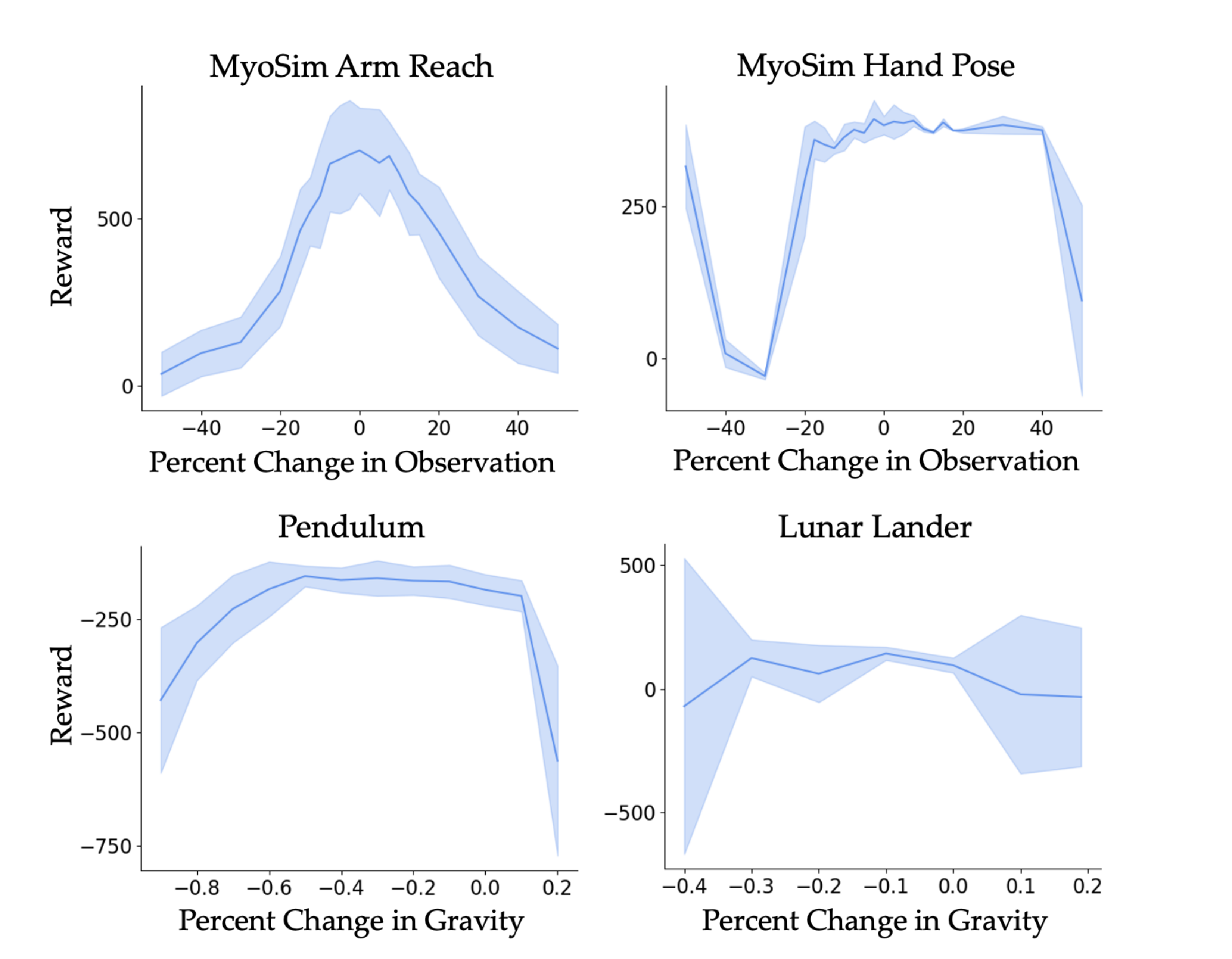}
	\caption{Evaluation reward when altering the environment during online training. We show that our approach is robust to a mismatch between simulation and reality.}
	\label{fig:modified_dynamics_results}
\end{figure}

We perform experiments shown in \cref{fig:modified_dynamics_results} to simulate the sim-to-real gap by systematically altering the dynamics of the environment during online training and testing while learning
$\convhat$. We demonstrate that our method is robust to these effects up to a point and that the online training aspect of our approach can help to account for a mismatch between simulated and real biomechanics. Specifically, we show that we can alter gravity by up to 40\% during online interaction in Pendulum and maintain performance. We show that our approach is robust to nearly 30\% change in gravity in LunarLander. In our Myosim environments, we systematically alter the observations during online training and testing and show that the Arm Reach task maintains performance for 10\% shift in observations and Hand Pose can handle nearly a 40\% change.

\end{document}